# Human-in-the-Loop Control of Objective Drift in LLM-Assisted Computer Science Education

Mark Dranias PhD[1][0000000346151069] and Adam Whitley PhD[2]

[1] Asheville Institute for Memory and Longevity. Mills River, NC USA
[2] University of North Carolina Asheville, Asheville, NC USA
`awhitley@unca.edu`

**Abstract.** Large language models (LLMs) are increasingly embedded in computer science education through AI-assisted programming tools, yet such workflows often exhibit objective drift, in which locally plausible outputs diverge from stated task specifications. Existing instructional responses frequently emphasize tool-specific prompting practices, limiting durability as AI platforms evolve. This paper adopts a human-centered stance, treating human-in-the-loop (HITL) control as a stable educational problem rather than a transitional step toward AI autonomy. Drawing on systems engineering and control-theoretic concepts, we frame objectives and world models as operational artifacts that students configure to stabilize AI-assisted work. We propose a pilot undergraduate CS laboratory curriculum that explicitly separates planning from execution and trains students to specify acceptance criteria and architectural constraints prior to code generation. In selected labs, the curriculum also introduces deliberate, concept-aligned drift to support diagnosis and recovery from specification violations. We report a sensitivity power analysis for a three-arm pilot design comparing unstructured AI use, structured planning, and structured planning with injected drift, establishing detectable effect sizes under realistic section-level constraints. The contribution is a theory-driven, methodologically explicit foundation for HITL pedagogy that renders control competencies teachable across evolving AI tools.

## 1 Introduction

Large language models (LLMs) are now routinely encountered in computer science education through AI-assisted programming. These systems generate code autoregressively via next-token prediction, conditioning each output on an evolving interaction history [1]. Extended AI-assisted workflows frequently exhibit objective drift: deviation from an explicitly stated task specification as small errors accumulate [2,3]. In educational settings where correctness and constraints are central, objective drift is both a usability and a pedagogical issue.

A second challenge is curricular durability. Instructional responses emphasizing tool-specific prompting techniques become obsolete as platforms evolve [4, 5, 6]. This curriculum is rooted in critiques of LLMs from autonomy-oriented AI research: LLMs do not optimize over task-related objective functions [7]. This exposes a durable role for humans interacting with LLMs by applying basic system engineering principles—requirements specification, constraint definition, and verification [8].

This paper adopts these constructs pedagogically, treating human-in-the-loop (HITL) control as a durable curricular design. The goal is for students to create explicit specifications that guide AI interactions regardless of which tool they use. This paper discusses the theory, outlines the pilot experimental study and presents a power analysis to estimate educational effects, rather than to provide confirmatory statistical evidence.

We propose a pilot undergraduate computer science laboratory curriculum that explicitly separates planning from execution. Students are trained to act as controllers by first creating (i) an objective specification of acceptance criteria and constraints, and (ii) a world model specification that binds the AI's option space (e.g., repository structure, permitted dependencies, architectural/algorithmic constraints). During execution, students use AI assistance to develop code within these bounds while monitoring for misalignment. In selected labs, the curriculum also deliberately injects objective drift that targets specific learning objectives to create structured opportunities for diagnosis and recovery. The pilot addresses two research questions:

**RQ1: AI-Assisted Coding Performance.** Does explicit training in objective specification and identifying world model constraints improve student performance in AI-assisted coding?

**RQ2: Concept Learning.** Can deliberate injection of objective drift on primary curricular topics improve student learning of specific computer science concepts?

We expect trained students will develop more sophisticated code artifacts faster than untrained students. With respect to RQ2, previous work suggests over-reliance on AI can impair learning [9]. We anticipate injecting topical drift will force students to pay more attention to key curricular concepts to offset that impairment.

## 2      Theory: HITL Control of Autoregressive Predictors

### 2.1      LLMs as World Models: Next-State Prediction and Option Space Control

A recent biologically motivated theory of AI autonomy proposes that a configurator selects task-relevant objectives while providing constraints on the option space of a world model [7]. World models are generally defined by the ability to predict next states from current ones, and while some feel LLMs make imperfect real-world models, few challenge their ability to model discrete closed systems relevant to computer programming [10]. When world models are viewed as discrete-state Markov processes, the set of possible next tokens represents the option space. We adopt this control-theoretic vocabulary as guidance in an educational HITL context.

### 2.2 Educational Goals

Our core premise is that effective AI-assisted programming depends on systems engineering and control competencies that are often implicit or underemphasized in traditional curricula: Objective articulation—specifying acceptance criteria and constraints precisely enough to support verification; Task and constraints definitions—making assumptions and boundaries explicit (repository structure, permitted dependencies, architectural intent); Monitoring and diagnosis—detecting deviations such as incorrect behavior, constraint violations, or uncontrolled scope expansion; Recovery—applying corrective actions that restore alignment (tightening constraints, revising objectives).

To emphasize these skills, the curriculum separates problem solving into planning and execution phases. The planning phase asks students to work with instructors and AI to analyze a programming problem into a set of objectives and constraints. The execution phase is done with an AI-assisted coding agent.

To help teach the competencies of diagnosis and recovery, our pedagogy also introduces deliberately injected objective drift: controlled perturbations that force students to reconcile outputs with specific requirements (e.g. off-by-one errors tied to indexing conventions). This allows instructors to target specific curricular concepts, trains HITL control skills, and makes control failure easily observable. Students are told they will occur and are expected to be vigilant and correct them.

From a learning-science perspective, the curriculum's core artifacts—objective specifications, world-model constraints, and specification-compliance monitoring—can be interpreted as externalized self-regulation supports. Explicit objective articulation corresponds to goal setting and planning; world-model constraints make learners' mental models of algorithms, invariants, and architectures inspectable; and execution-phase monitoring encourages metacognitive control and help-seeking behaviors [11]. This positions the intervention within established constructs of self-regulated learning and human–AI collaboration, while remaining distinct from prompt-training approaches that emphasize linguistic interaction strategies [12]. Prior work has shown that structured prompting and AI literacy can improve short-term performance [12], and that carefully designed errors can promote learning when they require active diagnosis [13]. Our contribution integrates these into a durable, tool-agnostic methodology in which curriculum-aligned errors are deliberately injected and learners are taught to detect, explain, and recover from misalignment. We treat human-in-the-loop control itself as the target competency.

## 3 Methods

### 3.1 Study Design Overview

We propose a pilot study in an undergraduate computer science laboratory course to evaluate a curriculum that frames AI-assisted programming as a HITL control problem. The study adopts a three-arm design intended to estimate educational benefits. Students are trained to construct and enforce control and system design abstractions—objective specifications and world-model constraints—when working with AI-assisted coding.

### 3.2   Participants and Setting

Participants are undergraduate students, and the study is embedded in the normal instructional setting using a standard IDE (e.g. Visual Studio Code) and an AI coding assistant. All participants complete the same programming tasks and assessments. Study arms divide course sections/lab groups to minimize contamination.

### 3.3   Experimental Conditions

The three study arms differ in instruction and workflow, not in the tasks themselves or AI tool access. Arm 1 serves as a baseline for unstructured AI use. Arm 2 isolates the effect of explicit HITL control abstractions. Arm 3 isolates the additional effect of deliberate objective drift on concept learning.

**Arm 1: Unstructured AI Use.** Students complete laboratory assignments using their usual programming workflow with unstructured AI assistance. No required planning artifacts are collected, and no constraints are imposed on interactions with AI systems.

**Arm 2: HITL Control Training.** Students complete assignments using a two-stage workflow that explicitly separates planning from execution.

In the planning stage, before any substantial AI code generation, students produce (i) an objective specification (e.g. quality measures, acceptance criteria checklists) and (ii) world model constraints (e.g. task scope, libraries, architectures, and permitted approaches). These artifacts are iteratively refined with instructor feedback and reflective AI assistance. The end product of this stage is a project instruction file that can be attached to a prompt or entered into the context of the LLM.

In the execution stage, students implement the assignment with AI assistance, within the bounds defined by the objective and world model. Students are instructed to monitor outputs relative to these artifacts, to treat violations as control failures, and make corrections to produce a final deliverable that satisfies them.

**Arm 3: HITL Control Training with Injected Drift.** Identical to Arm 2, except that during execution the AI is subjected to predefined drift perturbations at specified workflow landmarks to emphasize targeted curricular concepts. Students are informed the AI will make errors that they expected to detect and correct by relying on their planning documents and knowledge of programming principles. Each injected-drift event is specified in advance by: a curricular concept, a workflow landmark for injection, a fixed message appended to the AI context without student knowledge, and an expected failure signature observable in outputs. For example, to target Python's 0-indexing, the AI is biased toward one-based indexing at first appearance in the coding exercise, inducing a predictable off-by-one error output. These drift injections are limited in number, aligned with explicit learning objectives, and do not affect student grades.

### 3.4 Curriculum Sequence and Procedures

Participants in all 3 arms complete 2 instructional lab projects, followed by a common evaluation lab and a traditional assessment. Instructional labs are used to practice the assigned workflow (unstructured in Arm 1, HITL control in Arms 2 and 3) and, in Arm 3, to expose students to injected drift aligned with specific concepts. These labs are intended for practice and not for student evaluation; repeating the process twice should familiarize them with our HITL control training before the evaluation lab.

The evaluation lab is designed to address RQ1 and consists of a novel but related programming task completed under standardized conditions across all participants. Arms 2 and 3 will go through the same planning and execution phases as the instructional labs. No injected drift is used in the evaluation lab and all have AI access.

The traditional assessment aims to address RQ2 and assess any learning differences between arms. It consists of a traditional quiz or exam which includes questions on the same educational concepts from Arm 3's objective drift.

### 3.5 Data, Measures and Outcomes

Collected data include: Planning artifacts for Arms 2 and 3 on the instructional labs and evaluation lab; Repository histories and early versions of instructional labs obtained via version-control systems; De-identified interaction logs with the AI (e.g. prompts, AI calls); Final code submissions from the evaluation lab; De-identified, aggregated results from the traditional assessment. Interaction logs are used to characterize control behavior. In all data, personally identifying information is removed to preserve confidentiality.

**RQ1 (AI-assisted coding performance).** Here we address whether structured planning improves student performance at AI-assisted coding. The evaluation lab is the primary test and we measure: rubric-scored quality of objective and world model constraints (clarity, completeness, enforceability), specification compliance (required features, constraint adherence), frequency of constraint violations (e.g., unauthorized dependencies, file proliferation), and acceptance tests of the final code submission

**RQ2 (concept learning).** Here we address whether injected objective drift improved student understanding of specific educational concepts. During the traditional assessment, we introduce questions on these specific concepts and compare the aggregated performance of students in each arm of the study.

### 3.6 Statistical Analysis Plan

This pilot assesses feasibility and estimates effect sizes under section-level assignment; analyses therefore emphasize estimated effects and confidence intervals rather than causal inference. Primary outcomes are analyzed in a one-way linear model framework. Time-on-task will be an outcome; if incomplete, it will be summarized descriptively.

**RQ1 (AI-assisted coding performance)**. The primary outcome is evaluation lab performance. The primary planned one-sided contrast tests whether Arm 2 (HITL control) outperforms Arm 1 (unstructured AI use): $H_0: \mu_2 \leq \mu_1$ vs. $H_A: \mu_2 > \mu_1$. Arm 3 is included for secondary exploratory comparisons; omnibus ANOVA summaries are reported descriptively. Secondary outcomes (e.g. constraint violations, planning artifact quality) are summarized descriptively with exploratory contrasts to probe mechanisms.

**RQ2 (concept learning)**. Concept learning is assessed via targeted delayed-assessment items aligned to injected drift. The primary planned one-sided contrast tests whether Arm 3 (AI + injected drift) outperforms Arm 2 (AI without drift): $H_0: \mu_3 \leq \mu_2$ vs. $H_A: \mu_3 > \mu_2$. Additional descriptive comparisons (including Arm 1) and an omnibus ANOVA are reported to describe patterns between arms.

**Power/Sensitivity (Table 1).** A sensitivity analysis computes one-sided power ($\alpha = 0.05$) for the two planned contrasts (RQ1: $\mu_2 > \mu_1$; RQ2: $\mu_3 > \mu_2$) for arm sizes $n \in \{10, 15, 20\}$. Scores were assumed to be grades with an approximately normal distribution with mean 70-85 and a fixed SD $\sigma = 10$ for all arms.

A custom function was defined to compute power from three inputs: effect size d, sample size per arm n, and significance level α = 0.05. The function calculates the degrees of freedom (df = 2n - 2) and non-centrality parameter ($\lambda = d\sqrt{(n/2)}$), then returns the power using the non-central t-distribution (scipy.stats.nct).

Because students are assigned by section rather than individually, we adjusted for clustering using the design effect correction n_eff = n / [1 + (m-1)ρ], where m is section size and ρ = 0.10 is the intraclass correlation. Power ≥ 0.80 served as a cutoff.

## 4    Results

As summarized in Table 1, with per-arm sizes of 10–20 students, conventional power (80%) is achieved only for large effects, approximately 0.85–1.15 SD without clustering and ≥1 SD under plausible ICC; on a 100-point scale (σ = 10), this corresponds to differences of roughly 9–12 points, informing rubric resolution.

**Table 1.** Power/Sensitivity Analysis for RQ1 and RQ2

| Per-arm n | ICC ρ | n_eff | $d_{80}$ | $\Delta_{80}$ (points) |
|---|---|---|---|---|
| **10** | 0.00 | 10.00 | 1.15 | 11.5 |
| **10** | 0.10 | 5.26 | 1.45 | 14.5 |
| **15** | 0.00 | 15.00 | 1.00 | 10.0 |
| **15** | 0.10 | 8.11 | 1.25 | 12.5 |
| **20** | 0.00 | 20.00 | 0.85 | 8.5 |
| **20** | 0.10 | 11.11 | 1.05 | 10.5 |

*Note*. $d_{80}$ is min standardized mean difference required for 80% power; $\Delta_{80} = d_{80}\sigma$ assumes σ = 10 points on a 100-point scale. Effective sample size n_eff = n / [1 + (n− 1)ρ]

## 5  Discussion

A central tension in AI-assisted education is the tradeoff between short-term productivity and long-term conceptual learning. Unstructured AI use can produce correct artifacts while bypassing the cognitive work associated with a deep understanding of the problem, its solutions, and core programming concepts. This work proposes reframing AI-assisted programming as a system design and human-in-the-loop control problem, rather than merely desiring to elicit better code from increasingly capable models. This pedagogy is based on the assumption that system specification, diagnosis and stabilization are teachable skills. The pilot curriculum trains students to act as controllers who specify objectives, constrain the option space, and stabilize execution.

We introduce deliberate drift in pilot arm 3 based on observations that controlled perturbations can be instructionally valuable [13]. The injected-drift arm is designed to treat drift as a controlled instructional resource, analogous to designed errors or counterexamples in other instructional paradigms: it creates friction at conceptually meaningful points, prompting explanation, diagnosis, and repair.

A key design goal of the curriculum is durability under rapid tool change. Because objectives, constraints, and feedback loops are properties of tasks and workflows rather than of specific models or interfaces, the instructional content remains relevant even as AI systems evolve. This positions HITL control training as a stable target for Artificial Intelligence in Education (AIED) research, in contrast to curricula that hinge on transient prompt styles or platform-specific affordances [4, 5, 6].

Whether similar control abstractions can be translated to less structured and more open-ended domains remains an open empirical question.

### 5.1  Conclusion and Future Work

Grounding instructional design in system design and control abstractions rather than tool features, the curriculum offers a pathway for durable, theory-informed interventions that remain relevant despite rapid technological change. Future work includes developing libraries of concept-aligned drift injections, predictive rubrics for objective and world model quality, and transfer of HITL control training to domains other than software development with weaker feedback. These directions treat human agency not as a temporary workaround, but as a central design pillar in AI-assisted programming.

**Ethics and Responsible Use.** This study involves student participants and will obtain approval from the authors' institutional ethics review process. Participation in research data collection is voluntary and does not affect course grades; students may decline or withdraw without penalty. Students are informed that AI-assisted programming may produce unreliable outputs; in the injected-drift arm, students are informed curriculum-aligned errors are intentionally introduced to support diagnosis and learning, with a debrief after completion. Collected data is minimized, stored securely, and handled in accordance with institutional privacy policies. AI-assisted tools were used in manuscript preparation; the authors remain responsible for content.

**Disclosure of Interests.** The authors have no competing interests to declare relevant to the article.